\documentclass{article}
\usepackage{placeins}
\usepackage{amsmath}
\usepackage{amsfonts}
\usepackage{graphicx}
\usepackage{microtype}
\usepackage{gensymb}
\usepackage{enumitem}
\usepackage[affil-it]{authblk}
\usepackage[british]{babel}
\usepackage{csquotes}
\usepackage{xspace}
\newcommand{\etal}{\emph{et al.}\xspace}

\renewcommand{\thefootnote}{\fnsymbol{footnote}}
\title{TimeCluster with PCA is Equivalent to Subspace Identification of Linear Dynamical Systems}
\author{
  Christian L. Hines$^1$\thanks{Corresponding authors: \{chines, sspillard\}@turing.ac.uk} \quad Samuel T. Spillard$^1$$^*$ \quad Daniel P. Martin$^2$\footnote{The majority of this research was carried out while D. P. Martin was a member of The Alan Turing Institute.}\\
    \small{$^1$The Alan Turing Institute \quad $^2$Camulos}
}
\date{}

\begin{document}

\maketitle
\renewcommand{\thefootnote}{\arabic{footnote}}

\begin{abstract}
TimeCluster is a visual analytics technique for discovering structure in long multivariate time series by projecting overlapping windows of data into a low-dimensional space. We show that, when Principal Component Analysis (PCA) is chosen as the dimensionality reduction technique, this procedure is mathematically equivalent to classical linear subspace identification (block-Hankel matrix plus Singular Vector Decomposition (SVD)). In both approaches, the same low-dimensional linear subspace is extracted from the time series data. We first review the TimeCluster method and the theory of subspace system identification. Then we show that forming the sliding-window matrix of a time series yields a Hankel matrix, so applying PCA (via SVD) to this matrix recovers the same principal directions as subspace identification. Thus the cluster coordinates from TimeCluster coincide with the subspace identification methods. We present experiments on synthetic and real dynamical signals confirming that the two embeddings coincide. Finally, we explore and discuss future opportunities enabled by this equivalence, including forecasting from the identified state space, streaming/online extensions, incorporating and visualising external inputs and robust techniques for displaying underlying trends in corrupted data.
\end{abstract}

\section{Introduction}

\begin{figure}[ht]
\includegraphics[width=\textwidth]{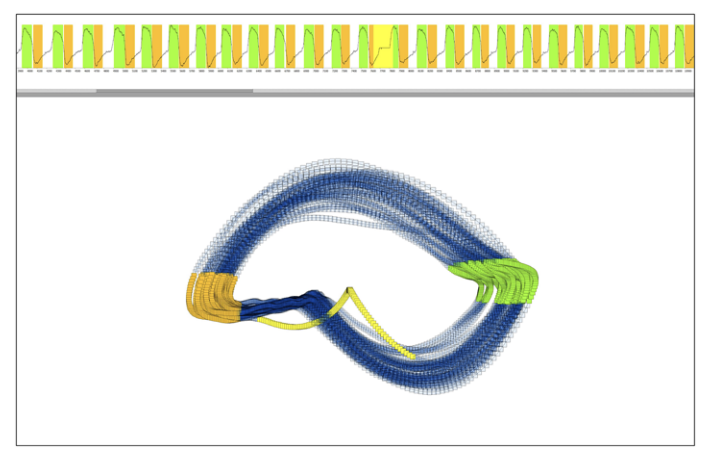}
\caption{An image of TimeCluster from~\cite{TimeCluster}. The periodicity of the time series can clearly be seen as a loop in the embedding. Highlighting similar data (orange and green) shows the repeating structure from the time series. The clear anomaly in the embedding (yellow) highlights a slightly subtle time series anomaly in the original data.}
\label{fig:timecluster}
\end{figure}

Long multivariate time series arise in domains as varied as environmental monitoring, industrial process control and human physiology, yet their intrinsic high dimensionality and complex temporal dependencies often obscure the patterns, correlations and anomalies that lie hidden within. Traditional line‑plot visualisations struggle to scale beyond a handful of variables or to highlight recurrent motifs over long horizons. Ali \etal proposed the TimeCluster procedure to address these challenges by first embedding each overlapping window of the multivariate signal into a high‑dimensional `trajectory' space and then applying dimensionality reduction (initially PCA) to project that trajectory into two dimensions~\cite{TimeCluster}. The resulting connected plot preserves both temporal ordering and similarity structure, enabling analysts to see repeated regimes, transitions and outliers at a glance, even in very long recordings. Figure~\ref{fig:timecluster} shows the original method in action.

Providing a rigorous mathematical foundation for the TimeCluster heuristic not only demystifies its empirical success but also unlocks principled extensions by situating the method within a well-understood theoretical framework. In doing so, we gain guarantees on consistency and enable the direct import of advances from system identification and statistical learning.

\section{Background}
We begin by providing all background information that is required to show an equivalence between TimeCluster and and Subspace Identification in Section~\ref{sec:tc_ss_equivalence}.

\subsection{TimeCluster Method}

In what follows we use lowercase letters for variables, bold letters for vectors and uppercase letters for matrices. Given a total number of time steps $T$, let $\boldsymbol{y}_t$ be a $D$-dimensional vector at time $t$. The full multivariate time series is then represented as $Y_{1:T} \in \mathbb{R}^{T \times D}$.

\paragraph{Sliding‑Window Embedding}

First, apply min–max scaling independently to each variate so that its values lie in $[0,1]$. Next, choose a window length $L$ (with stride $s=1$). For a multivariate time series $Y_{1:T}$, construct the `trajectory' matrix:
$$
Z \in \mathbb{R}^{(T - L + 1)\times LD}.
$$
This construction treats each of the $T - L + 1$ rows of $Z$ as a time point in $\mathbb{R}^{LD}$ using  interleaving of the $D$ channels.

\paragraph{Embedding Example}

Suppose we have a two-channel ($D=2$) time series with $T=4$ steps:
\begin{equation}\label{eq:signal-example}
    Y_{1:4} = 
    \begin{bmatrix}
        1 & 10 \\
        2 & 20 \\
        3 & 30 \\
        4 & 40
    \end{bmatrix}
\end{equation}
Taking the window length $L=2$ results in $T-L+1 = 3$ windows. Form the trajectory matrix by taking each 2‑step window as a \emph{row} (channel‑interleaved):
$$
Z = 
\begin{bmatrix}
y_{1,1} & y_{2,1} & y_{1,2} & y_{2,2} \\[4pt]
y_{1,2} & y_{2,2} & y_{1,3} & y_{2,3} \\[4pt]
y_{1,3} & y_{2,3} & y_{1,4} & y_{2,4}
\end{bmatrix}
=
\begin{bmatrix}
1 & 10 & 2 & 20 \\[4pt]
2 & 20 & 3 & 30 \\[4pt]
3 & 30 & 4 & 40
\end{bmatrix}\in\mathbb R^{3\times 4}.
$$

\paragraph{Dimensionality Reduction}

In the original TimeCluster system, linear PCA is one option among others (t-SNE, UMAP, or convolutional autoencoders for feature extraction), but the essential idea is this sliding-window embedding followed by dimensionality reduction. Given the matrix $Z$ (whose rows are observations), PCA computes the eigenvectors of the covariance matrix or equivalently performs an SVD on $Z$, thereby projecting the high-dimensional observations into a low-dimensional space for visualisation while preserving most of the structure. The top $r$ principal components are the first $r$ eigenvectors (or singular vectors) corresponding to the largest eigenvalues, and they provide the best rank-$r$ approximation to $Z$ in the least-squares sense. Projecting the data onto the first two components yields a 2D scatter plot.

\paragraph{Visualising the Results}
Ali \etal then plot these embeddings in an interactive plot with the original time series plotted above the embedding. Selecting points in the embedding highlights points in the time series, and vice versa, allowing easy exploration. The resulting tool also demonstrates some nice properties:

\begin{itemize}
    \item Periodicity in the time series forms loops in the embedding.
    \item Anomalies in the time series cause anomalies in the embedding.
    \item Changepoints in the time series cause clusters in the embedding.
\end{itemize}

\begin{figure}[ht]
\includegraphics[width=\textwidth]{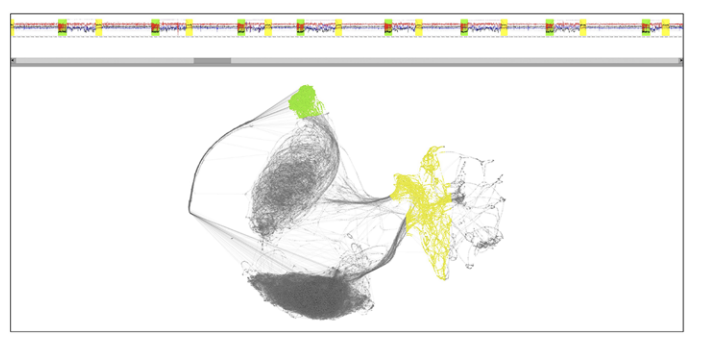}
\caption{An image of TimeCluster from~\cite{TimeCluster}. Clusters in the embedding space correspond to self similar sections of the time series. Two examples (green and yellow) are highlighted to demonstrate.}
\label{fig:timecluster_cluster}
\end{figure}

Figure~\ref{fig:timecluster} demonstrates the impact of periodicity and anomalies. Figure~\ref{fig:timecluster_cluster} demonstrates the relationship of changepoints and clustering.

\subsection{Linear Dynamical Systems}

A linear time-invariant (LTI) state-space model with (hidden) state $x_t\in\mathbb{R}^n$, time-dependent input $u_t\in\mathbb{R}^m$ and observed output $y_t\in\mathbb{R}^D$ can be written as 
\begin{subequations}\label{eq:lti-system}
    \begin{align}
        \boldsymbol{x}_{t+1} &= A \boldsymbol{x}_t + B \boldsymbol{u}_t + \boldsymbol{w}_t, \\
        \boldsymbol{y}_t &= C \boldsymbol{x}_t + D \boldsymbol{u}_t + \boldsymbol{v}_t.
    \end{align}
\end{subequations}
Here $\boldsymbol{w}_t,\boldsymbol{v}_t$ are process and measurement noise terms and:
\begin{description}[nosep]
    \item $A\in\mathbb{R}^{n\times n}$ determines how the latent state evolves over time
    \item $B\in\mathbb{R}^{n\times m}$ determines how the input influences the hidden state
    \item $C\in\mathbb{R}^{D\times n}$ determines how the latent state is measured
    \item $D\in\mathbb{R}^{D\times m}$ determines how the input influences the observations
\end{description}
Standard conditions (controllability, observability) ensure the state can be uniquely identified from output data~\cite[Definitions 7.1 and 8.1]{Aastroem2021}. Many commonly used time series models, such as ARIMA processes, admit an equivalent state-space form of this type~\cite{Durbin2012}. Once $(A,B,C,D)$ and the state sequence are known, one can simulate or forecast future outputs by iterating the state equation.

\subsection{Subspace Identification}

Subspace identification (ID) methods aim to estimate the state-space model parameters $(A,B,C,D)$ from input-output data. Importantly, subspace ID algorithms do not require iterative optimisation, avoiding issues of local minima that affect techniques such as the expectation–maximisation (EM) algorithm; instead, all steps reduce to linear algebra operations (for a more complete overview see~\cite{Verhaegen1992,VanOverschee1996}).

We start by noting that the LTI system in equation~(\ref{eq:lti-system}) can be formulated as a matrix equation
\begin{equation}\label{eq:least-squares-form}
    \begin{bmatrix}
        \boldsymbol{x}_{t+1} \\
        \boldsymbol{y}_t
    \end{bmatrix}
    =
    \begin{bmatrix}
        A & B \\
        C & D
    \end{bmatrix}
    \begin{bmatrix}
        \boldsymbol{x}_t \\
        \boldsymbol{u}_t
    \end{bmatrix}
    +
    \begin{bmatrix}
        \boldsymbol{w}_t \\
        \boldsymbol{v}_t
    \end{bmatrix}.
\end{equation}
Given we have access to inputs $\boldsymbol{u}_t$ and outputs $\boldsymbol{y}_t$, an estimate of the hidden state vector $\boldsymbol{x}_t$ would allow the system parameters
\begin{equation}
    \Theta :=
    \begin{bmatrix}
        A & B \\
        C & D
    \end{bmatrix}
\end{equation}
to be estimated using standard least squares. 

\paragraph{Estimating the hidden state vector}

The recursive nature of the LTI system allows observations \emph{after} time $t$ to be written in terms of the state vector \emph{at} time $t$, plus input and noise terms. For example, expanding $\boldsymbol{x}_{t+1}$ in the equation for the observation at time $t+1$:
\begin{align*}
    \boldsymbol{y}_{t+1} 
    &= C \boldsymbol{x}_{t+1} + D \boldsymbol{u}_{t+1} + \boldsymbol{v}_{t+1} \\
    &= C \left(A \boldsymbol{x}_t + B \boldsymbol{u}_t + \boldsymbol{w}_t\right) + D \boldsymbol{u}_{t+1} + \boldsymbol{v}_{t+1} \\
    &= C A \boldsymbol{x}_t
        + \underbrace{C B \boldsymbol{u}_t + D \boldsymbol{u}_{t+1}}_\text{input term} 
        + \underbrace{C \boldsymbol{w}_t + \boldsymbol{v}_{t+1}}_\text{noise term}.
\end{align*}
We will assume hereafter that the system is `output-only,' meaning $u_t = 0$ for $t=1,2,\dots,T$, so the input term vanishes. For input-output systems, algorithms like N4SID and MOESP use orthogonal projection to eliminate the contribution from the input (for more details see~\cite{Jamaludin2013,VanOverschee1994}).

Repeating the same approach for $L$ successive time steps, we arrive at the matrix equation
\begin{equation}\label{eq:recursive-per-timestep}
    \begin{bmatrix}
        \boldsymbol{y}_t \\
        \boldsymbol{y}_{t+1} \\
        \vdots \\
        \boldsymbol{y}_{t+L-1}
    \end{bmatrix} 
    =
    \underbrace{
    \begin{bmatrix}
        C \\
        CA \\
        \vdots \\
        CA^{L-1}
    \end{bmatrix}
    }_{\mathcal{O}_L}
    \boldsymbol{x}_{t} 
    + \epsilon,
\end{equation}
for some noise term $\epsilon$, and where we have defined the \emph{extended observability matrix} $\mathcal{O}_L\in\mathbb{R}^{LD \times n}$.

We can construct a system of equations like (\ref{eq:recursive-per-timestep}) for each time step. Note, however, that we cannot use all $T$ time steps; the lookahead window of length $L$ reduces the maximum number of equations to $T - L + 1$. Stacking each equation into columns yields the following matrix equation:
\begin{equation}
    H = \mathcal{O}_L\,X_{1:T-L+1} +\epsilon.
\end{equation}
Here, we have introduced the \emph{block-Hankel} matrix
\begin{equation}
    H = 
    \begin{bmatrix}
        \boldsymbol{y}_1 & \boldsymbol{y}_2 & \cdots & \boldsymbol{y}_{T-L+1} \\
        \boldsymbol{y}_2 & \boldsymbol{y}_3 & \cdots & \boldsymbol{y}_{T-L+2} \\
        \vdots & \vdots & & \vdots \\
        \boldsymbol{y}_L & \boldsymbol{y}_{L+1} & \cdots & \boldsymbol{y}_T
    \end{bmatrix}
    \in \mathbb{R}^{LD\times (T-L+1)}.
\end{equation}
and the state sequence matrix $X_{1:T-L+1} = [\boldsymbol{x}_1, \boldsymbol{x}_2, \ldots, \boldsymbol{x}_{T-L+1}] \in \mathbb{R}^{n \times (T-L+1)}$.

Classical results show that if $L$ is chosen such that the system order $n \leq L$, then $\text{rank}(H) = n$ in the noise-free case, and the SVD of $H$ yields the column space of the observability matrix and the row space of the controllability matrix~\cite{Ho1966,Gopinath1969}. Concretely, one computes $$H = U\Sigma V^T$$ and takes the first $n$ columns $U_{1:n}$ of $U$ (and corresponding singular values $\Sigma_{1:n}$). If, as in the majority of cases, $n$ is not known \emph{a priori}, then one can estimate the system order by taking the top $r$ singular values above a predetermined threshold $\varepsilon$. If $r = n$, these columns span the same space as the true state sequence. The state estimates can then be computed as $\hat{X}_{1:T-L+1} = \Sigma_{1:n} V_{1:n}^T$ (where $V_{1:n}$ contains the first $n$ columns of $V$), and one recovers $(A, B, C, D)$ by least-squares fitting of the original system in equation~(\ref{eq:least-squares-form}).

\paragraph{Block Hankel Example}

Consider the same signal as in equation~(\ref{eq:signal-example}), the block‑Hankel matrix is built by stacking two `lagged' block‑rows
\begin{equation*}
    H = 
    \begin{bmatrix}
        \boldsymbol{y}_1 & \boldsymbol{y}_2 & \boldsymbol{y}_3\\[6pt]
        \boldsymbol{y}_2 & \boldsymbol{y}_3 & \boldsymbol{y}_4 
    \end{bmatrix}
    = 
    \begin{bmatrix} 
        \;[1,\,10] & [2,\,20] & [3,\,30]\\[6pt]
        \;[2,\,20] & [3,\,30] & [4,\,40] 
    \end{bmatrix}.
\end{equation*}
If you `flatten' each $[a,b]$ into two rows, the result is:
\begin{equation*}
    H =
    \begin{bmatrix} 
    1 & 2 & 3\\ 
    10 & 20 & 30\\ 
    2 & 3 & 4\\ 
    20 & 30 & 40 
    \end{bmatrix}\in\mathbb{R}^{4\times 3}.
\end{equation*}

\section{Equivalence of TimeCluster and Subspace ID}
\label{sec:tc_ss_equivalence}

We now formalise why TimeCluster with PCA is equivalent to subspace identification. Form the block-Hankel matrix $H\in\mathbb{R}^{LD\times (T-L+1)}$ as above. Observe that each column of $H$ is a contiguous window $[y_t,\dots,y_{t+L-1}]^\top$. TimeCluster instead forms a data matrix $Z\in\mathbb{R}^{(T-L+1)\times LD}$ whose \emph{rows} are these windows. Consequently, we see that TimeCluster's data matrix is the transpose of the block-Hankel matrix: 
\begin{equation}
    Z = H^\top.
\end{equation}
Performing PCA on the rows of $Z$ is thus equivalent to computing the SVD of $H$. Indeed, if 
\begin{equation}
    H = U \Sigma V^\top
\end{equation}
is the SVD of the block-Hankel matrix, then 
\begin{equation}
    Z = H^\top = V \Sigma U^\top.
\end{equation}
This reveals the fundamental equivalence: Subspace ID computes SVD of $H$ to get $U_n$ (observability subspace) and state sequence $\hat{X}_{1:T-L+1} = \Sigma_{1:n} V_{1:n}^T$. Meanwhile, TimeCluster-PCA computes SVD of $Z$ to get $V_{1:n}$ (principal components) and coordinates $Z V_{1:n} = V \Sigma_{1:n}$.

Both yield equivalent low-dimensional coordinates for the temporal windows. The only difference is in their interpretation: subspace ID views the coordinates as hidden states of a dynamical system, while TimeCluster views them as principal component coordinates for visualisation and clustering. This shows that the fundamental process behind TimeCluster is a principled modelling technique. Thus, embeddings created through TimeCluster faithfully represent the underlying state trajectory of these time series, lending credence to the theory that anomalous features in the TimeCluster embedding space represent anomalous sections of the time series being considered, and are not just an artefact of the TimeCluster alogithm. 

\paragraph{TimeCluster's stride parameter}

The $Z$ matrix construction used by TimeCluster~\cite{TimeCluster} is fundamentally equivalent to the block Hankel matrix when the stride parameter $s = 1$, differing only in orientation (rows vs. columns). When $s > 1$, the stride parameter functions as a downsampling preprocessing step, in the columns of the Hankel matrix (striding over rows in the TimeCluster matrix). However, it is worth noting that this resulting matrix is no longer a Hankel matrix; it contains less information that the full Hankel matrix, but more information than the Hankel matrix on the time series downsampled by $s$. From here on, we consider only the $s=1$ case, in order to be consistent with Subspace Identification methods.

\section{Numerical Experiments}
In what follows we are comparing the mathematical details of the two models, not the user experience. By showing the equivalence, it means our mathematical guarantees and additional properties can be added to their nice GUI, rather than us constructing a brand new tool.
\subsection{Synthetic Data}

\begin{figure}[ht]
\includegraphics[width=\textwidth]{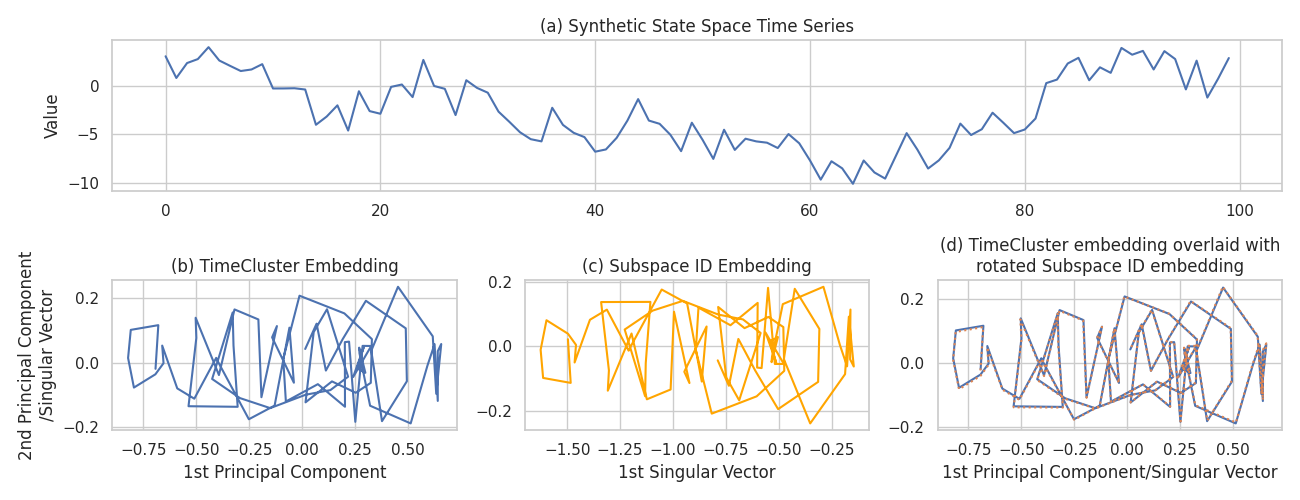}
\caption{Side-by-side comparison of embeddings from TimeCluster and Subspace ID method on synthetic AR(2) data.}
\label{fig:ar2_tc_ss}
\end{figure}

For our first numerical experiment, we set up a contrived example where we generate an AR(2) process with known state-space dimension $n=2$. From here, we run the original TimeCluster algorithm over the series, with a sliding window of length $L=2$, stride $s=1$, and PCA with $r=2$ output components. We also apply the subspace ID method to the same sequence, and obtain the first $n=2$ left singular vectors $U_{1:2}$.

Whilst they are not the most informative, the resultant embeddings from both methods are shown side-by-side in Figure~\ref{fig:ar2_tc_ss} - where it easy to see that they result in the same output, up to a rotation and translation factor (in this case, we shift the subspace ID embeddings along the x-axis such that the coordinates are aligned in the x-direction, and also mirror the embeddings in the x-axis), as shown in Figure~\ref{fig:ar2_tc_ss}(d).

\subsection{Real-World Data}

\begin{figure}[ht]
\includegraphics[width=\textwidth]{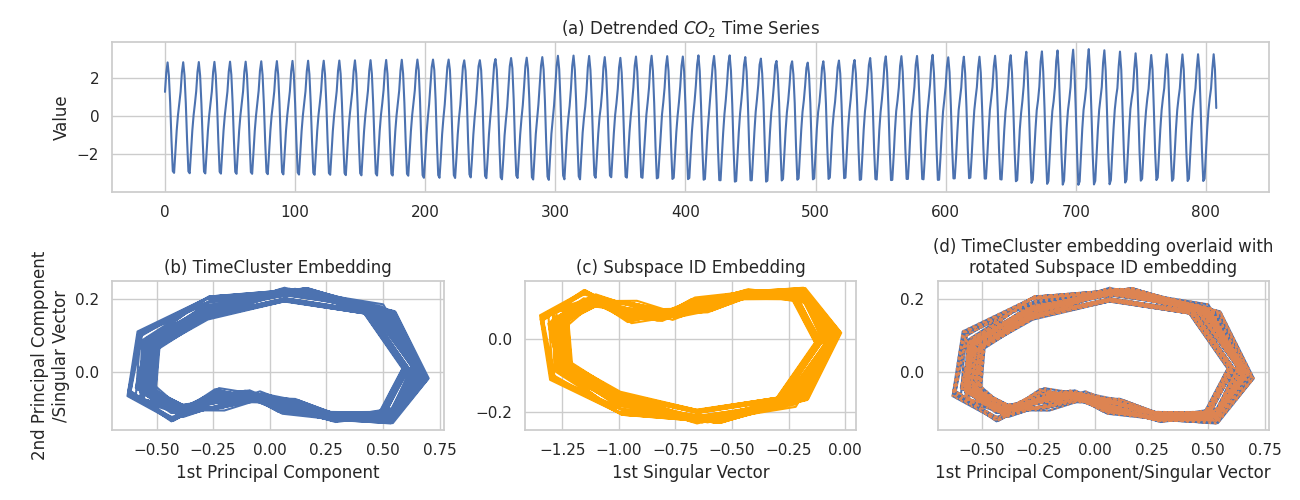}
\caption{Side-by-side comparison of embeddings from TimeCluster and Subspace ID method on real $CO_2$ data.}
\label{fig:co2}
\end{figure}

For our second numerical experiment, we use the well-known publicly available Mauna Loa $CO_2$ dataset~\cite{MaunaLoa}. We normalise it by removing the trend, leaving us with just the seasonal component. Once again, we apply both TimeCluster PCA and Subspace ID embeddings to the data, and show that the resultant figures line up identically in Figure~\ref{fig:co2} (this time, mirrored in the x-axis, as well as some translation along the x-axis).

\section{Extensions Enabled by Subspace ID}

The equivalence between TimeCluster and subspace ID invites several possible extensions.

\subsection{Extended Visualisation}
\begin{figure}[ht]
\includegraphics[width=\textwidth]{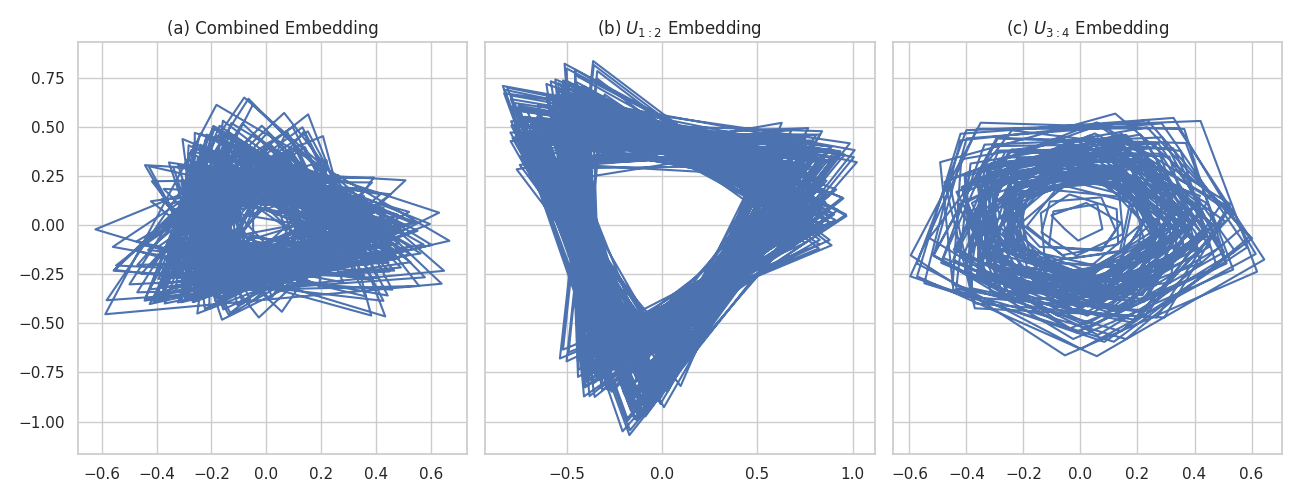}
\caption{Side-by-side comparison of embeddings from TimeCluster on a time series with double periodicity. Being able to determine that the state space model of this time series has dimension $n=4$ means we can plot each of the two periodicities independently. More generally any insights a user has into the state space structure can now be pulled out into the visualisation.}
\label{fig:2periodicity}
\end{figure}

Now that our embeddings are meaningful due to coming from a state space model, we can extend our visualisation, using additional knowledge if required. For example, if our time series has two periodicities (e.g. a frequency of $\frac{1}{3}$ and $\frac{1}{5}$, as in Fig.~\ref{fig:2periodicity}), these will be reflected in the state. It is then possible to plot these two components separately, on two embedding plots, to aid the user in understanding their data.

\subsection{Forecasting}
As shown in Section~\ref{sec:tc_ss_equivalence}, the fact that TimeCluster and Subspace ID produce the same embeddings but with different interpretations, implicitly means that the TimeCluster PCA algorithm effectively recovers a predictive state-space model. Viewing the TimeCluster embedding as $\hat{X}^{T}_{1:T-L+1}$, we then need only estimate:

\begin{align*}
    \text{State transition matrix}&: A \\
    \text{Process noise covariance}&: Q \\
    \text{Observation matrix}&: C \\
    \text{Observation noise covariance}&: R
\end{align*}
in order to use our TimeCluster embeddings to create predictions using the Kalman forward equations~\cite{welch1995introduction}. 

We can estimate $\hat{A}$ via least squares using the following:

$$
\hat{A} = \hat{X}_{1:\tau} \hat{X}_{0:\tau-1}^{\dagger}
$$

We can estimate $\hat{C}$, also using least squares, as:

$$
\hat{C} = Y\hat{X}^{\dagger}
$$

We can then estimate both $Q$ and $R$ emprically using the following:

\begin{align*}
    \boldsymbol{w}_t &= \boldsymbol{x}_{t+1} - \hat{A}\boldsymbol{x}_t \\
    \boldsymbol{v}_t &= \boldsymbol{y}_t - \hat{C}\boldsymbol{x}_t \\
    \hat{Q} &= \frac{1}{T - 1}\sum^{T-1}_{t=1}\boldsymbol{w}_t\boldsymbol{w}_{t}^{\top} \\
    \hat{R} &= \frac{1}{T}\sum^{T}_{t=1}\boldsymbol{v}_t\boldsymbol{v}_{t}^{\top}
\end{align*}

Finally, we can use our estimated $\hat{A}$, $\hat{C}$, $\hat{Q}$ and $\hat{R}$ (along with an uninformative prior for $P_t$ of the identity, $I$), to propagate our state-space model forward in time recursively, like so:

\begin{align*}
    \hat{\boldsymbol{x}}_{t|t-1} &= A\hat{\boldsymbol{x}}_{t-1|t-1} \\
    P_{t|t-1} &= AP_{t-1|t-1}A^{\top} + Q \\
    K_t &= P_{t|t-1}C^{\top}(CP_{t|t-1}C^{\top} + R)^{-1} \\
    \hat{\boldsymbol{x}}_{t|t} &= \hat{\boldsymbol{x}}_{t|t-1} + K_t (\boldsymbol{y}_t - C\hat{\boldsymbol{x}}_{t|t-1}) \\
    P_{t|t} &= (I - K_tC)P_{t|t-1}
\end{align*}

where $K_k$ is the Kalman gain, and $(\boldsymbol{y}_k - C\hat{\boldsymbol{x}}_{k|k-1})$ is the residual between our predicted observation, and the actual observation. Note that we are using a fully on-line model here, where our state and covariance estimates are updated by new observations before estimating the next observation.

\begin{figure}[ht]
\includegraphics[width=\textwidth]{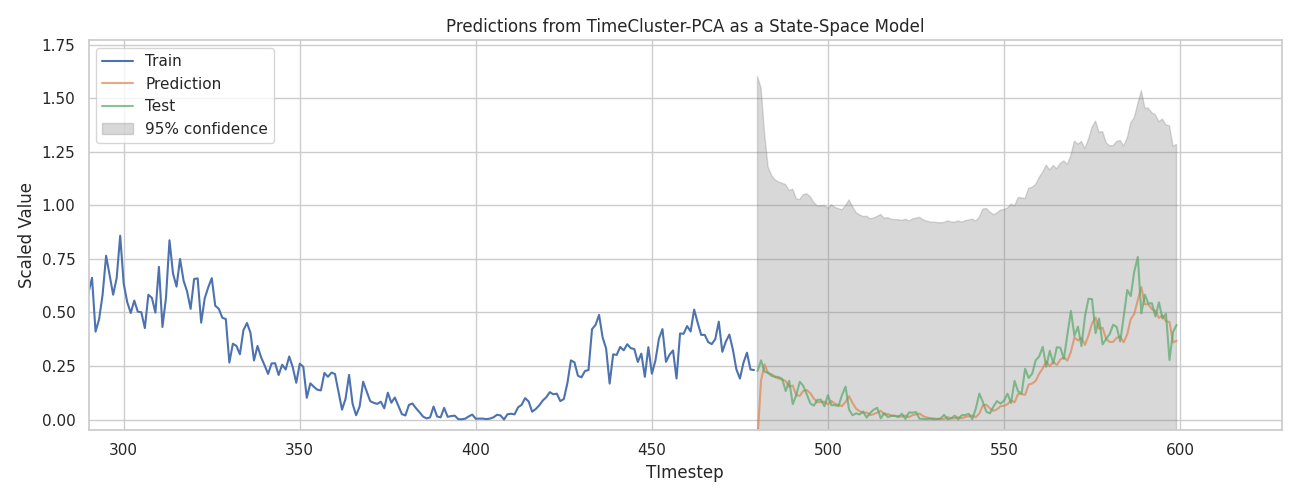}
\caption{Kalman forward predictions of sunspot dataset using TimeCluster embeddings to estimate the state-space model.}
\label{fig:sunspot_predict}
\end{figure}

In Figure~\ref{fig:sunspot_predict}, we show the predictions from this model on the publicly available monthly mean sunspot dataset~\cite{SILSO_Sunspot_Number}. Here, we have taken the embeddings from the TimeCluster algorithm and used them to derive the required parameters for our state-space model, then propagated predictions forwards using the Kalman forward equations, updating our matrix estimates as new observations are provided.

The advantage of this formalisation enabling forecasting is around user interpretability. If a user notes that all points of interest fall within a small area of space within the TimeCluster projection, they can ask ``When is this time series next expected to enter this region of space?". This can potentially be much easier for them to ask than when a particular shape in the time series is expected to appear again.

\subsection{Exogenous Regressors}

Viewing the TimeCluster-PCA algorithm as a subspace ID problem operating on the output-only block Hankel matrix, i.e. $\boldsymbol{u}_t=0$ for $t=1,\dots,T$, allows us to extend TimeCluster into scenarios where there are external regressors. In this setting, we can instead operate on an input-output Hankel matrix in order to calculate our embeddings. By embedding these time-dependent `control' inputs, $\boldsymbol{u}_t$, alongside our outputs, we can more easily visualise how the system is driven. 

\begin{figure}[ht]
\includegraphics[width=\textwidth]{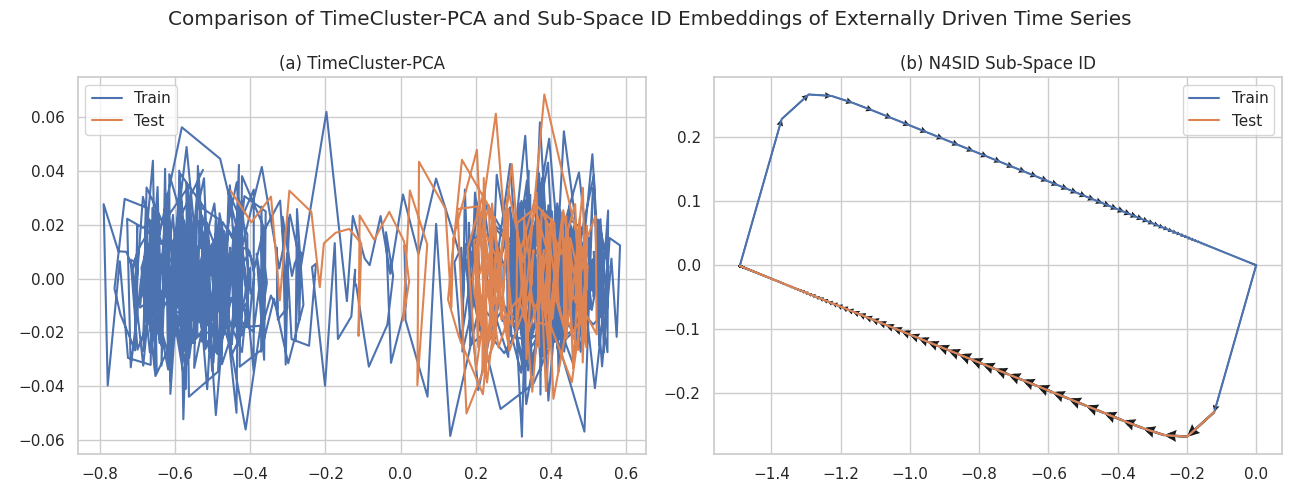}
\caption{Comparison of (a) TimeCluster-PCA embeddings and (b) subspace ID embeddings generated using N4SID algorithm. Blue and orange sections correspond to equivalent sections of the time series in Figure~\ref{fig:exog_synth}. Note the orange line showing a transition from one region to another, as the regressor drives the system, which is easier to isolate in the subspace ID embeddings.}
\label{fig:exog_embed}
\end{figure}

For this example, we generate data from an $n=2$ state-space model, driven by a stepped exogenous regressor. The resultant time series is shown in Figure~\ref{fig:exog_synth}. We then use the \verb|NFourSID| package~\cite{NFourSID} to model this dataset using the N4SID algorithm. The package itself takes care of performing subspace identification with exogenous regressors, but we are left with (among other things) estimates for the state trajectory, which we can use as coordinates for the visualisation. When comparing the traditional TimeCluster-PCA embeddings with this N4SID embedding in Figure~\ref{fig:exog_embed}, we can see that the N4SID method provides a noise-free visualisation using just the learned state trajectories, as opposed to the TimeCluster-PCA alogrithm whose embeddings are susceptible to the noise in the observations.

\subsection{Kalman Smoother for Visual Denoising}

\begin{figure}[ht]
\includegraphics[width=\textwidth]{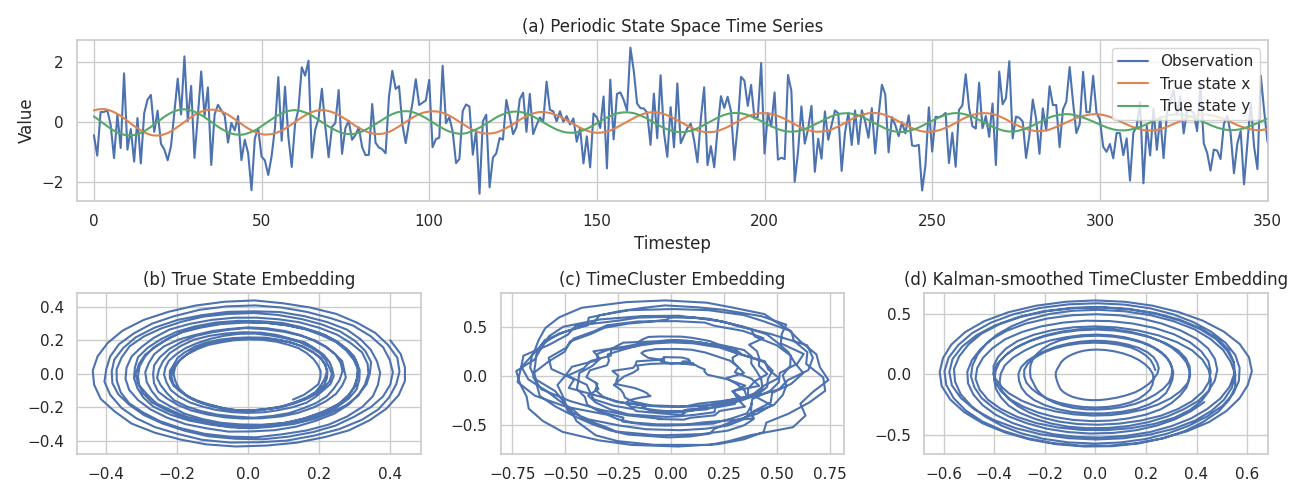}
\caption{(a) Synthetic periodic time series derived from state-space model, plotted with true hidden states. Comparison of (b) true state trajectory embeddings, (c) TimeCluster-PCA embeddings and (d) TimeCluster-PCA embeddings after Kalman smoother applied.}
\label{fig:synth_periodic}
\end{figure}

As can be seen in some earlier figures, the estimated state trajectories from the Hankel SVD approach can be very noisy. Whilst the dimensionality reduction step aims to keep only those singular vectors that contribute to the signal (rather than the noise), it is a purely empirical technique that can still be susceptible to noise. If we instead combine our estimated subspace identification parameters with a Kalman smoother, we can refine our state trajectory estimates, often leading to a more smooth, noise-free state trajectory. In Figure~\ref{fig:synth_periodic}(a) we show a synthetic periodic state-space time series. Figure~\ref{fig:synth_periodic} also shows the state embedding for (b) the true state trajectory, (c) the TimeCluster-PCA state trajectory and (d) the TimeCluster-PCA state trajectory after smoothing with a Kalman smoother. One can see the Kalman-smoothed TimeCluster-PCA embeddings are closer to the true state trajectory embeddings, and more visually clean.

\subsection{Additional Further Extensions}

\paragraph{Online/streaming methods} By using incremental SVD or PCA algorithms, one could update the Hankel subspace in real time as new data arrive. Efficient online algorithms for updating a low-rank SVD exist, enabling a streaming variant of TimeCluster for live data~\cite{Cardot2018}.

\paragraph{Robust subspace methods} Real data often contain outliers or missing values. Robust PCA techniques, e.g. low-rank plus sparse decompositions, can identify a low-rank Hankel subspace in the presence of gross corruption~\cite{Candes2011}. Similarly, one can use structured matrix completion to handle missing observations. Incorporating these ideas would make TimeCluster more robust to noise and data gaps.

\section{Conclusion}

We have shown that the TimeCluster method with PCA is equivalent to classical linear subspace identification: the sliding-window embedding yields a Hankel matrix whose SVD (via PCA) produces exactly the same subspace as subspace ID algorithms. This establishes a direct link between time series visual analytics and control-oriented system identification. In practice, one can therefore leverage system-theoretic tools (model estimation, Kalman filtering, modal analysis) within the TimeCluster framework and vice versa. We suggest a melding of these two domains into a single product that offers visualisation, model estimation, and intuitive forecasting, would be an extremely useful tool for analysts looking to triage very long time series datasets.

\clearpage
\bibliographystyle{ieeetr}
\bibliography{refs}

\begin{thebibliography}{10}

\bibitem{TimeCluster}
M.~Ali, M.~W. Jones, X.~Xie, and M.~Williams, ``{TimeCluster},'' vol.~35, no.~6, pp.~1013--1026.

\bibitem{Aastroem2021}
K.~J. Åström and R.~Murray, {\em Feedback Systems}.
\newblock Princeton University Press, 2~ed.

\bibitem{Durbin2012}
J.~Durbin and S.~J. Koopman, {\em Time Series Analysis by State Space Methods}.
\newblock No.~38 in Oxford Statistical Science Series, Oxford University Press, 2~ed.

\bibitem{Verhaegen1992}
M.~Verhaegen and P.~Dewilde, ``Subspace model identification part 1,'' vol.~56, no.~5, pp.~1187--1210.

\bibitem{VanOverschee1996}
P.~Van~Overschee and B.~De~Moor, {\em Subspace Identification for Linear Systems}.
\newblock Springer New York.

\bibitem{Jamaludin2013}
I.~W. Jamaludin, N.~A. Wahab, N.~S. Khalid, S.~Sahlan, Z.~Ibrahim, and M.~F. Rahmat, ``{N4SID} and {MOESP} subspace identification methods,'' in {\em 9th International Colloquium on Signal Processing and its Applications}, pp.~140--145, IEEE.

\bibitem{VanOverschee1994}
P.~Van~Overschee and B.~De~Moor, ``{N4SID},'' vol.~30, no.~1, pp.~75--93.

\bibitem{Ho1966}
B.~L. Ho and R.~E. Kalman, ``Effective construction of linear state-variable models from input/output functions,'' vol.~14, no.~1–12, pp.~545--548.

\bibitem{Gopinath1969}
B.~Gopinath, ``On the identification of linear time-invariant systems from input-output data,'' vol.~48, no.~5, pp.~1101--1113.

\bibitem{MaunaLoa}
X.~Lan and R.~Keeling, ``Trends in co2.''

\bibitem{welch1995introduction}
G.~Welch, G.~Bishop, {\em et~al.}, ``An introduction to the kalman filter,'' 1995.

\bibitem{SILSO_Sunspot_Number}
F.~{Clette} and L.~{Lefèvre}, ``Silso sunspot number v2.0.'' https://doi.org/10.24414/qnza-ac80, 07 2015.
\newblock Published by WDC SILSO - Royal Observatory of Belgium (ROB).

\bibitem{NFourSID}
{S. van Gemert}, ``Nfoursid.''

\bibitem{Cardot2018}
H.~Cardot and D.~Degras, ``Online principal component analysis in high dimension,'' vol.~86, no.~1, pp.~29--50.

\bibitem{Candes2011}
E.~J. Candès, X.~Li, Y.~Ma, and J.~Wright, ``Robust principal component analysis?,'' vol.~58, no.~3, pp.~1--37.

\end{thebibliography}
\clearpage
\appendix
\section{Synthetic Dataset with Exogenous Regressor}
\begin{figure}[!h]
\includegraphics[width=\textwidth]{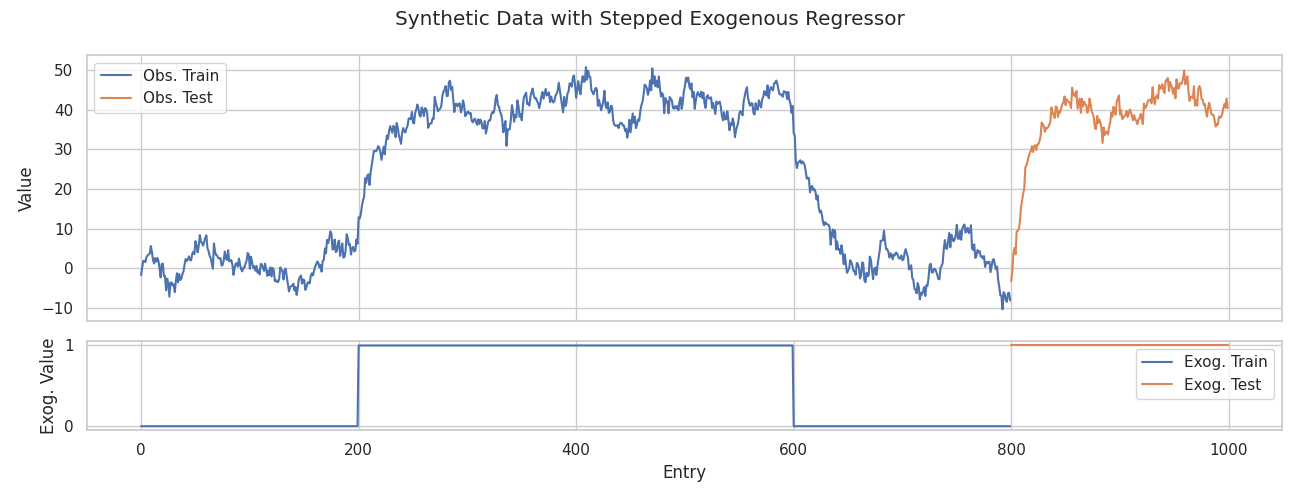}
\caption{Synthetic dataset driven by a stepped exogenous regressor. Split 80:20 into a train and test set for purposes of visualising how the embeddings are driven by the exogenous regressor.}
\label{fig:exog_synth}
\end{figure}
\FloatBarrier

\end{document}